\pdfoutput=1

\documentclass[11pt]{article}

\usepackage[final]{acl}

\usepackage{times}
\usepackage{latexsym}

\usepackage[T1]{fontenc}

\usepackage[utf8]{inputenc}

\usepackage{microtype}

\usepackage{inconsolata}

\usepackage{graphicx}
\usepackage{arydshln}

\title{OnlySportsLM: Optimizing Sports-Domain Language Models with SOTA Performance under Billion Parameters}

\author{
 Zexin Chen\\
 New York University\\
 \texttt{zc2404@nyu.edu} \\
 \And
 Chengxi Li\\
 Carnegie Mellon University\\
 \texttt{chengxil@andrew.cmu.edu} \\
 \AND
 Xiangyu Xie\\
 Cornell University\\
\texttt{xx358@cornell.edu}\\
 \And 
 Parijat Dube\\
 IBM Research\\
 \texttt{pdube@us.ibm.com}
}

\begin{document}
\maketitle{}
\begin{abstract}
This paper explores the potential of a small, domain-specific language model trained exclusively on sports-related data. We investigate whether extensive training data with specially designed small model structures can overcome model size constraints. The study introduces the \texttt{OnlySports} collection, comprising \texttt{OnlySportsLM}, \texttt{OnlySports Dataset}, and \texttt{OnlySports Benchmark}. Our approach involves: 1) creating a massive 600 billion tokens \texttt{OnlySports Dataset} from FineWeb, 2) optimizing the RWKV-v6 architecture for sports-related tasks, resulting in a 196M parameters model with 20-layer, 640-dimension structure, 3) training the \texttt{OnlySportsLM} on part of \texttt{OnlySports Dataset}, and 4) testing the resultant model on \texttt{OnlySports Benchmark}. \texttt{OnlySportsLM} achieves a 37.62\%/34.08\% accuracy improvements over previous 135M/360M state-of-the-art models and matches the performance of larger models such as SomlLM 1.7B and Qwen 1.5B in the sports domain. Additionally, the \texttt{OnlySports} collection presents a comprehensive workflow for building high-quality, domain-specific language models, providing a replicable blueprint for efficient AI development across various specialized fields.

\end{abstract}

\section{Introduction}
General-purpose large language models (LLMs) have demonstrated remarkable capabilities across various tasks~\cite{minaee2024largelanguagemodelssurvey}.  However, such performance comes at the cost of excessive computational resources and sometimes inefficiencies in domain-specific applications. Domain-specific language models offer a promising alternative, potentially achieving comparable or superior performance in targeted areas while significantly reducing model size. 
\\

\noindent Despite their potential, recent domain-specific models face several challenges. Large models such as BloombergGPT \citep{wu2023bloomberggptlargelanguagemodel}, while powerful, requires extensive computational resources (e.g., 64 × 8 A100 40GB with a total of 1.3 million GPU hours), making them infeasible for most research institutions. Additionally, many domain models suffer from a lack of high-quality domain-specific text data, with models like BioMedLM \cite{bolton2024biomedlm27bparameterlanguage} trained on only 34.6 billion tokens and SportsBert \cite{SportsBERT} on merely 1-2 billion tokens. Furthermore, most domain models follow the model structure of general models, leaving room for optimization, especially for smaller model sizes.
\\

\noindent In light of these challenges, recent research on small general-purpose language models, such as MobileLLM \cite{liu2024mobilellmoptimizingsubbillionparameter} and SmolLM \cite{benallal2024smollm}, has provided valuable insights into efficient model structures. However, their effectiveness in domain-specific modeling remains unproven. To address these challenges and leverage recent insights, we propose a new approach for small domain-specific language models, utilizing specialized model structures and a collection pipeline for large in-domain corpus for efficient and cost-effective training. 
\\

\noindent To verify this approach, we choose sports as the target domain due to its unique combination of broad public interest, rich content, and a constant influx of new data through ongoing events and competitions. Moreover, sports language often contains domain-specific jargon, statistics, and contextual nuances that general-purpose models may struggle to capture accurately. By focusing on sports, we can demonstrate the potential of domain-specific models in a field that is both widely accessible and technically challenging. Additionally, the sports domain provides an excellent testbed for evaluating a model's ability to handle real-time information processing and generation, skills that are crucial in many real-world applications. Based on this approach and domain selection, we present \texttt{OnlySports\footnote{Our collection is available at: \url{https://huggingface.co/collections/Chrisneverdie/onlysports-66b3e5cf595eb81220cc27a6}}}, a novel framework for developing high-performance, small-scale sports language models.
\subsection{Contributions}


1. \textbf{\texttt{OnlySports Dataset}}: A large-scale, high-quality sports-specific text corpus of 600 billion tokens, extracted from the FineWeb dataset \cite{penedo2024finewebdatasetsdecantingweb}.

2. \textbf{\texttt{OnlySports Benchmark}}: A novel evaluation method for assessing sports knowledge generation, using 1000 diverse prompts and state-of-the-art (SOTA) language models for evaluation.

3. \textbf{\texttt{OnlySportsLM}}: A 196 million parameter RWKV-v6\footnote{Our training code is available at: \url{https://github.com/BlinkDL/RWKV-LM}} \cite{peng2024eaglefinchrwkvmatrixvalued} based sports language model trained on half of the OnlySports Dataset. In our OnlySports Benchmark, OnlySportsLM outperforms the preceding SOTA general purpose 135M/360M language model by 37.62\%/34.08\%.

\section{Collection of Domain Data}
In this section, we present the path to building \texttt{OnlySports Dataset}, a comprehensive collection of English sports documents. This dataset comprises a diverse range of content including news articles, blogs, match reports, interviews, and tutorials, all extracted from the FineWeb dataset. FineWeb is a thoroughly cleaned and deduplicated subset of CommonCrawl, spanning from 2013 to present. It represents one of the best open-source datasets for LLM training. Our extraction process involved two key steps: first, we applied URL filtering to identify potentially relevant content, and second, we developed a custom sports text classifier to accurately identify and extract sports-related documents from the filtered data. The resulting \texttt{OnlySports Dataset} encompasses 1.2 TB of text, equivalent to approximately 600 billion RWKV tokens. This makes it the largest sport domain dataset to date, significantly surpassing previous collections in both scale and comprehensiveness.
\subsection{URL Filtering}
To efficiently identify potentially sports-related content within the FineWeb dataset, we implemented a preliminary URL filtering step. We carefully select a list of sports-related terms, encompassing various sports, leagues, brands, and media. This approach allows us to rapidly narrow down the dataset to documents likely to contain sports content.
\\

\noindent Our keywords include:
\begin{itemize}
    \item General sports terms: \textit{sport, athletic, athlete, fitness, workout, gym, league, team, champion, football, soccer, basketball, baseball, tennis, cricket, rugby, golf, volleyball, hockey, cycling, swimming, wrestling, running, boxing, racing, swim, goal}
    \item Major leagues and organizations: \textit{NFL, NBA, MLB, NHL, FIFA, UEFA, NCAA, MMA, UFC, WWE, Premier League, LaLiga, Bundesliga, SerieA, Ligue1, EPL, NASCAR, MotoGP, Formula1, F1}
    \item Sports events, brands, and media: \textit{Olympic, cup, playoff, marathon, copa, Nike, Adidas, ESPN, BleacherReport, SI.com, news}

\end{itemize}
\noindent We applied these keywords in both their standard and capitalized forms where appropriate (e.g., NBA/nba, FIFA/fifa). This keyword list ensured a high recall in identifying potential sports content, which was then further refined by our classification model. Although the list does not exhaustively cover all sports, the nature of sports websites often includes the word \textit{sport} in their URL, ensuring broad coverage of sports-related content.
\begin{table}[h!]
\centering
\resizebox{\linewidth}{!}{%
\begin{tabular}{lcccc}
\hline
\textbf{Class} & \textbf{Precision} & \textbf{Recall} & \textbf{F1-Score} & \textbf{Support} \\ \hline
0              & 0.98               & 0.98            & 0.98              & 3631             \\
1              & 0.99               & 0.99            & 0.99              & 6429             \\ \hline
\textbf{Accuracy}     & & &0.99 &10060           \\
\textbf{Macro Avg}    & 0.99               & 0.99            & 0.99              & 10060           \\
\textbf{Weighted Avg} & 0.99               & 0.99            & 0.99              & 10060           \\ \hline
\end{tabular}%
}
\caption{Sports text classifier performance in the test set, correctly classifying most labels}
\label{tab:classifier-results}
\end{table}
\subsection{Sports Text Classifier}
To develop our sports text classifier, we first created a balanced dataset of sports and non-sports content. We manually scraped 64k samples from seven prestigious sports websites, selected to cover a wide range of sports topics. To balance this, we classified 36k non-sports text documents from a subset of FineWeb using GPT-3.5, ensuring diversity in the non-sports content. We then labeled this combined dataset, designating sports-related text as class 1 and non-sports text as class 0.
\\

\noindent  For the classification model, we chose \textit{Snowflake-arctic-embed-xs} \cite{merrick2024embeddingclusteringdataimprove} as our base due to its efficient performance on text classification tasks. We then add a binary classification layer to this model and train it for 10 epochs with a learning rate of 3e-4.
\\

\noindent Table \ref{tab:classifier-results} presents the performance metrics of our classifier, demonstrating its exceptional accuracy in distinguishing between sports and non-sports documents. The model achieves near-perfect precision, recall, and F1-scores for both classes, with an overall accuracy of 0.99.

\subsection{Data Filtering and Conversion}
\begin{figure}[t]
  \includegraphics[width=\columnwidth]{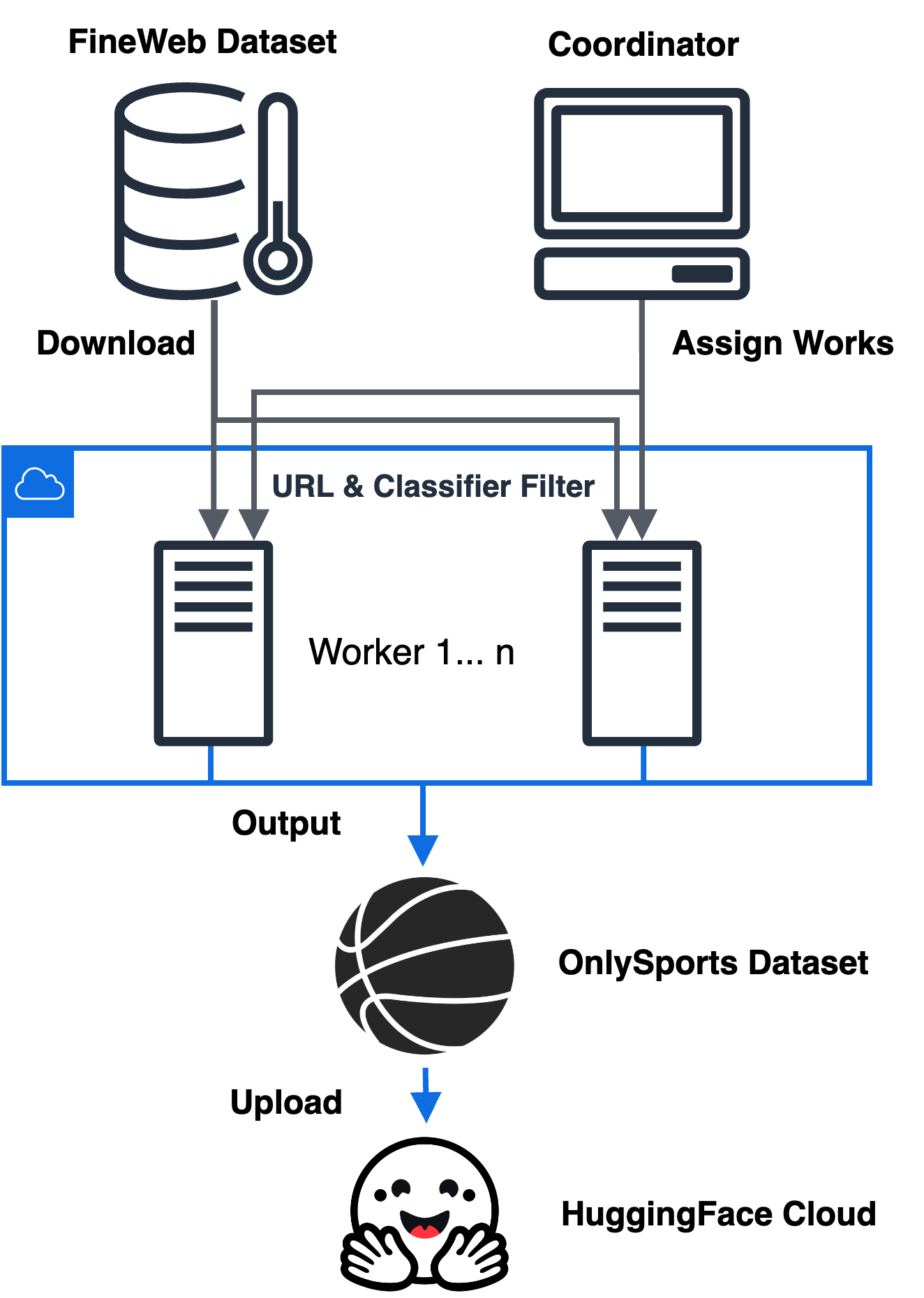}
  \caption {Data pipeline to create \texttt{OnlySports Dataset}}
  \label{graph:pipeline}
\end{figure}
Figure \ref{graph:pipeline} presents a scalable MapReduce architecture \cite{MapReduce} to filter sports-related content from the 90TB FineWeb dataset for model training. This approach allows us to overcome limitations in CPU resources and disk space.
\\

\noindent In the map phase, we use a Golang-based coordinator with the Gin Web framework to distribute tasks across eight Python-powered worker servers. The filtering process occurs in two steps: 1. URL keyword filtering, which reduced the dataset size by 85\%.
2. Application of our sports text classifier for further curation.
\\

\noindent The resulting filtered data is stored in parquet format and uploaded to HuggingFace. For the reduce phase, we utilized a high-capacity cloud server to tokenize the parquet files using an open-source Rust script. This streamlined pipeline enabled us to efficiently process the massive FineWeb dataset, extracting a high-quality sports-specific corpus for training \texttt{OnlySportsLM.}

\section{Optimizing Model Structure for Sports Domain}

We explore the potential for model structural optimization before training with the \texttt{OnlySports Dataset}. A previous study \citep{liu2024mobilellmoptimizingsubbillionparameter} suggests that general-purpose sub-billion parameter models perform better when using more layers than the traditional 12-layer design while having less dimensions. We hypothesize that domain-specific small models would also follow this deep and thin rule. We explore models with approximately 190M parameters and find results that partially support this principle

\subsection{Training Setup}
Our experiments are conducted on 8 H100 GPUs. We perform exploratory experiments on a 4.5B tokens subset of \texttt{OnlySports Dataset}. 
\\

\noindent We evaluated the pre-trained model on zero-shot commonsense reasoning tasks, including ARC-easy, ARC-challenge \cite{clark2018thinksolvedquestionanswering}, PIQA \cite{bisk2019piqareasoningphysicalcommonsense}, HellaSwag \cite{zellers2019hellaswagmachinereallyfinish}, as well as sports text generation task using \texttt{OnlySports Benchmark}.
\subsection{OnlySports Benchmark}
\label{sec:incomplete-sentence-generation}

We introduce a novel evaluation method inspired by the Hellaswag benchmark but targeted specifically for sports knowledge generation. Instead of asking multiple choice questions, our benchmark directly assesses a model's ability to complete sports-related prompts without fine-tuning, providing insight into sports-specific language understanding and generation capabilities. To ensure a comprehensive and relatively unbiased assessment, we employ multiple state-of-the-art language models as evaluators, assessing generated responses across two key criteria: accuracy and factuality, and continuity and relevancy. This approach allows for an evaluation of sports-related text generation capabilities across various models.

\subsubsection{Tag and Partial Sentence Generation}
\label{sub:tag}
To construct our evaluation dataset, we generated 50 diverse sports-related tags encompassing popular sports, major leagues, prominent athletes, and game strategies using GPT-4 API. These tags serve as the foundation for creating a comprehensive set of prompts. For each tag, we craft 20 incomplete sentences, resulting in a total of 1,000 prompts. Each prompt is intentionally designed to end abruptly, providing an ideal context for models to complete. The prompts incorporate well-known sports facts, statistics, or narratives, allowing assessment of a wide range of sports-related knowledge and generation capabilities. For instance, for the tag \textit{\#BasketballTeams}, the following partial sentence prompt is generated: \textit{Spurred on by the superstar duo of Shaquille O'Neal and Kobe Bryant, the L.A Lakers clinched three consecutive}. This abrupt ending sets the stage for models to complete the narrative. A well-trained model would likely continue the sentence with "\textit{NBA championships from 2000 to 2002}" or a similar factual completion, demonstrating its ability to maintain contextual coherence and accuracy.
\subsubsection{Model Inference and Evaluation Using SOTA LLMs}

In our inference process, each prompt is separately fed to the models. We employed consistent hyperparameter settings across all models, with temperature set to 1 and top-p value to 0.3, to ensure the generation of consistent, high-probability outputs. Each response is limited to 80 tokens.
\\
\begin{figure*}[h]
  \includegraphics[width=0.48\linewidth]{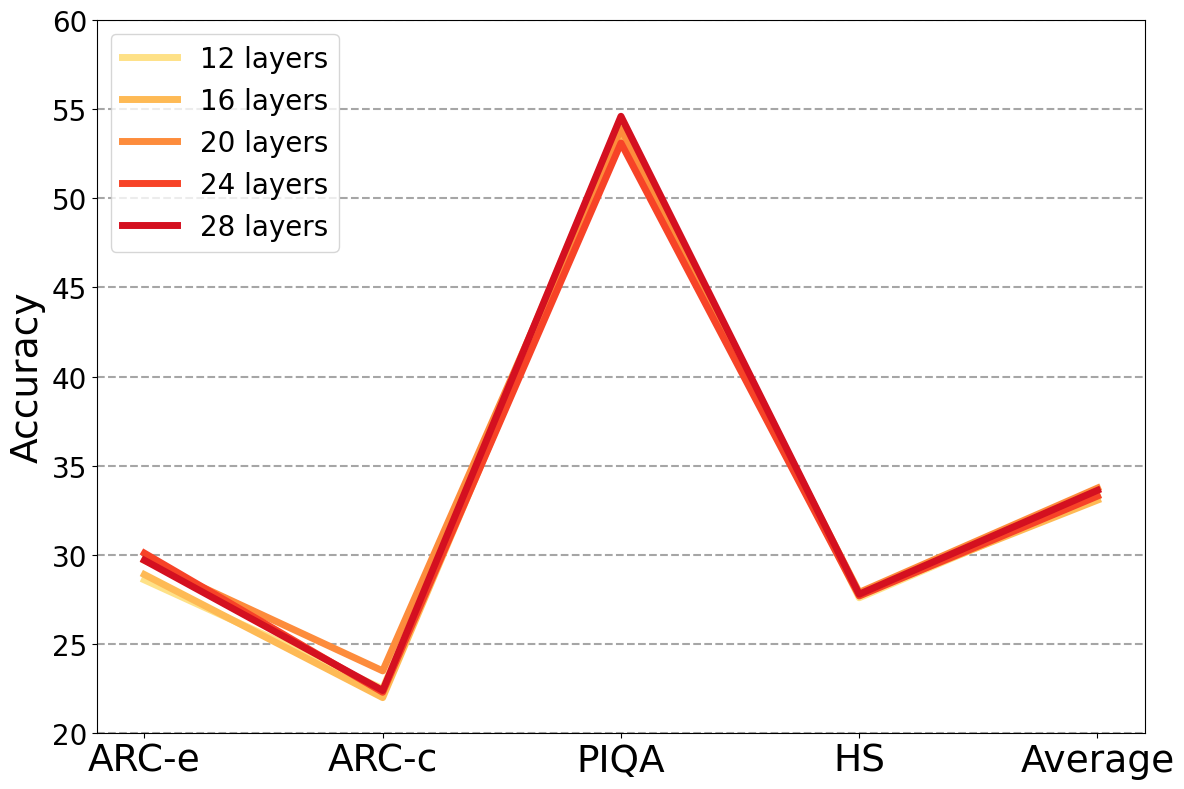} \hfill
  \includegraphics[width=0.48\linewidth]{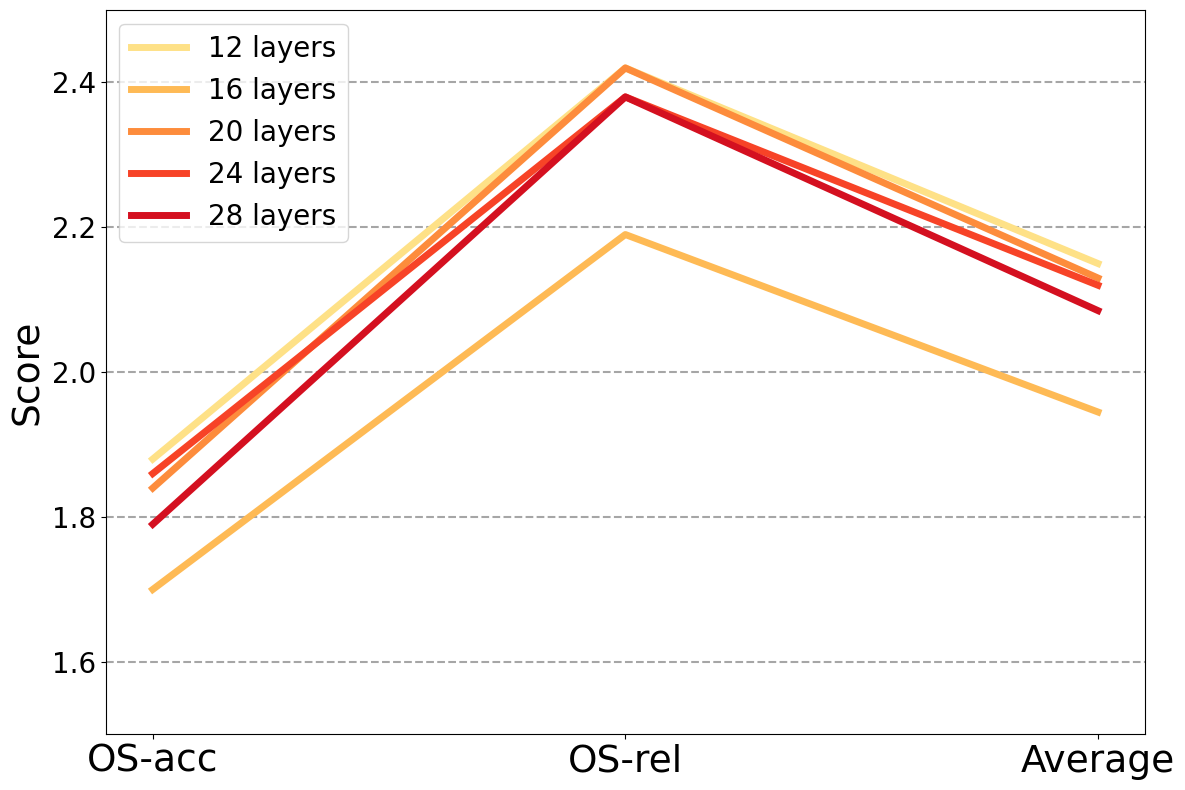}
  \caption {Performance comparison with varying depths and widths on \texttt{OnlySports Benchmark} and general zero-shot
evaluations}
\end{figure*}
\label{sec:text-evaluation}
\\
\noindent To evaluate the model-generated responses, we adopt an approach inspired by LLM-as-a-judge \cite{Zheng2023}, which approximates human preferences in assessing open-ended text. We utilize two state-of-the-art language models, GPT-4o and Claude 3.5 Sonnet, as evaluators. The assessment is conducted across two distinct criteria at a scale of 1-5, adhering to the principle of multi-dimensional evaluation as recommended by \citet{Zheng2023}.
To mitigate potential biases inherent in large language model judges, we implement several measures:
1. Deployment of multiple LLM judges to enhance reliability and reduce individual model biases.
2. Standardization of prompts and evaluation criteria to ensure consistency across assessments.
After scores are generated by each model, we take the average of them to be the final score.
\\

\noindent The input prompt format for evaluation is defined as follows:
\begin{itemize}
\item \textit{prompt: (partial sentence fed to the models)}
\item \textit{response: [SEP] Answer1 [SEP] Answer2 [SEP] Answer3...}
\end{itemize}
Where \textit{[SEP]} is a separator token used to distinguish between different model responses.
\\

\noindent The two evaluation criteria are defined as follows:
\begin{itemize}
    \item \textbf{Accuracy and Factuality:} Evaluates the model’s ability to generate accurate and fact-based continuations, ensuring that the information aligns with well-known sports facts and data. The score is denoted as \texttt{OS-acc} on a scale from 1 (mostly inaccurate with significant factual errors) to 5 (fully accurate and factually impeccable).
    
    \item \textbf{Continuity and Relevancy:} Assesses the relevance of the generated text to the given prompt, ensuring that the continuation is contextually appropriate and directly related to the previous sentence. This criterion, denoted as \texttt{OS-rel}, is scored from 1 (poor continuation that diverges significantly from the prompt's context) to 5 (excellent continuation that seamlessly extends the prompt's narrative, context, and style).
\end{itemize}

\noindent For each criterion, a system message with a detailed grading rubric is provided in the appendix for reference.

\begin{table*}[h]
\centering
\begin{tabular}{ccc|ccccccc}
\hline
\hline
\textbf{\#Layer} & \textbf{\#Dim} & \textbf{\#Param} & \textbf{final loss} & \textbf{OS-acc} & \textbf{OS-rel} & \textbf{ARC-e} & \textbf{ARC-c} & \textbf{PIQA} & \textbf{HS}\\
\hline
12 & 768 & 196M & 2.344 & \textbf{1.88} & \textbf{2.42} & 28.6 & 22.5 & 54.5 & 27.6\\
16 & 704 & 200M & 2.360 & 1.70 & 2.19 & 28.9 & 22.0 & 53.6 & 27.8\\
20 & 640 & 196M & \textbf{2.335} & 1.84 & \textbf{2.42} & 29.7 & \textbf{23.5} & 53.9 & \textbf{27.9}\\
24 & 576 & 185M & 2.338 & 1.86 & 2.38 & \textbf{30.1} & 22.3 & 53.1 & 27.7\\
28 & 512 & 169M & 2.364 & 1.79 & 2.38 & 29.7 & 22.4 & \textbf{54.6} & 27.8 \\
\hline
\hline
\end{tabular}
\caption{Model performance across varying architectures. Compares models with different layer counts and dimensions on \texttt{OnlySports Benchmark} and general zero-shot tasks (ARC-e, ARC-c, PIQA, Hellaswag).}
\label{tab:model_configs}
\end{table*}

\subsection{Depth and Width Experiments}
Our experimental results presented in Table \ref{tab:model_configs} reveal interesting insights about the relationship between model depth and performance. We conduct a study involving models ranging from 169M to 200M parameters, varying in depth from 12 to 28 layers and width from 512 to 768 dimensions. We observe that both traditional wider models and moderately deeper architectures perform well on \texttt{OnlySports Benchmark}. While the 12-layer wider model has the highest \texttt{OS-acc} (1.88) and \texttt{OS-rel} (2.42) scores, the 20 layers model shows comparable results in relevancy score and slightly less \texttt{OS-acc} (1.84). This finding, contrary to conclusion by \citet{liu2024mobilellmoptimizingsubbillionparameter} and \citet{benallal2024smollm}, underscores the need for task-specific architectural experimentation.
\\

\noindent General zero-shot tasks exhibit some benefits from increased depth, though less pronounced than in previous studies on general-purpose models. Models with 20 to 28 layers often outperform shallower configurations across various reasoning tasks.
\\

\noindent Based on these findings, we selected the L20D640 (20 layers, 640 dimensions) model for further training, balancing strong performance across domain-specific and general tasks. We denote this model as \texttt{OnlySportsLM}

\section{Experiments}
\subsection{Experimental Settings}
\begin{figure*}[t]
  \includegraphics[width=0.96\linewidth]{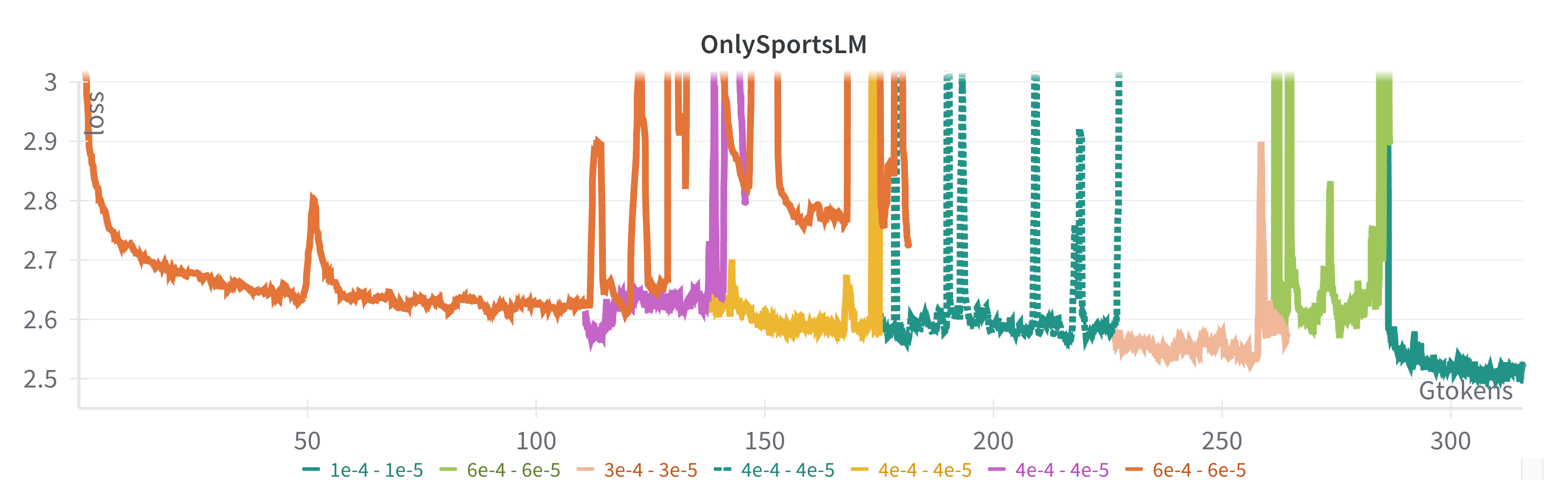}
  \caption {\texttt{OnlySportsLM} training loss over time with varying learning rates. The graph shows how loss fluctuates as we adjust the learning rate, starting from higher rates and gradually decreasing to stabilize training and reduce loss spikes. This insight is shared by the author of RWKV \citet{peng2024eaglefinchrwkvmatrixvalued}}
\label{fig:loss}
\end{figure*}

We train \texttt{OnlySportsLM} from scratch utilizing the AdamW optimizer \cite{loshchilov2019decoupledweightdecayregularization}  with a weight decay of 0.1 and a context length of 1024 tokens. Our experiments are performed on a cluster of 8 H100 GPUs, with a per-GPU batch size of 40. Following a cosine decay schedule, the initial learning rate is set to 6e-4. However, due to observed loss spikes during training, the learning rate is subsequently adjusted, ultimately being reduced to 1e-4 (detailed in Figure \ref{fig:loss}).  Due to constraints on available funding, the training stopped at 315B tokens in 7500 steps, which is around half the size of \texttt{OnlySports Dataset}.

\subsection{Main Results}
We compare the final \texttt{OnlySportsLM} checkpoint on \texttt{OnlySports Benchmark} and zero-shot commonsense reasoning tasks (Hellaswag, PIQA, ARC-challenge, and ARC-easy) with previous training checkpoints and recent open-source models. To ensure consistency in evaluation procedures, all models were assessed using their publicly available implementations from the HuggingFace model repository. General benchmark scores are retrieved from their corresponding paper.

\subsubsection{Sports Domain Generation}
Table \ref{tab:model_scores} compares our \texttt{OnlySportsLM} and two recent state-of-the-art general-purpose models, ranging from 137M to 1.7B parameters. We focused on two sets of models:
1. The SmolLM series \cite{benallal2024smollm}, with 137M, 360M, and 1.7B parameter models, reportedly surpasses the performance of all comparable small language models on general benchmarks. 
2. The Qwen2 collection \cite{yang2024qwen2technicalreport}, with 500M and 1.5B parameter models, also claims top performance on major benchmarks, even though they were trained on multilingual datasets.
These model collections,  released in June 2024 and July 2024 respectively, represent the latest development in small model research. For models under 1B parameters, \texttt{OnlySportsLM} outperforms all models by a significant margin. Notably, our model gains 34.44\% accuracy over Qwen2-0.5B while being 61\% smaller in size. Moreover, even when comparing to models over 1B parameter, our model performs only slightly worse (-5.23\%) than Qwen2-1.5B and marginally better (0.40\%) than SmolLM-1.7B in average score. This is a surprising result considering our model is only 12\% the size of SmolLM-1.7B.

\begin{table*}[h]
\centering
\begin{tabular}{lc|ccccccc}
\hline
\hline
\setlength\dashlinedash{0.2pt}
\setlength\dashlinegap{0.5pt}
\textbf{Model} & \textbf{\#Params} & \textbf{OS-acc} & \textbf{OS-rel} & \textbf{OS-Avg.} & \textbf{ARC-e} & \textbf{ARC-c} & \textbf{PIQA} & \textbf{HS}\\
\hline
\multicolumn{9}{c}{\textit{number of parameters $<$ 1B}} \\
\hdashline[1pt/1pt]
OnlySportsLM & 196M & \textbf{2.157} & \textbf{2.847} & \textbf{2.502} & 37.2 & 23.5& 59.6& 37.8\\
SmolLM-135M & 135M & 1.684 & 1.951 & 1.818 &  43.9 & - & 69.9 & 42.3\\
SmolLM-360M & 360M & 1.705 & 2.027 & 1.866 &  \textbf{51.1} & - &\textbf{72.0} & \textbf{53.8}\\
Qwen2-0.5B & 500M & 1.645 & 2.077 & 1.861 & 39.7 & \textbf{31.5} & 69.3 & 49.3\\
\hline
\multicolumn{9}{c}{\textit{number of parameters $\geq$ 1B}} \\
\hdashline[1pt/1pt]
Qwen2-1.5B & 1.5B & \textbf{2.327} & \textbf{2.952} & \textbf{2.640} & 48.2 & \textbf{43.9} & 75.5 & \textbf{66.6}\\
SmolLM-1.7B & 1.7B & 2.261 & 2.723 & 2.492 & \textbf{61.5} & - & \textbf{77.3} & 64.1\\
\hline
\hline

\end{tabular}
\caption{Performance comparison of \texttt{OnlySportsLM} against state-of-the-art models. Our model outperforms larger sub-1B models on sports tasks and competes with 1B+ models, raw scores provided in Appendix \ref{sub:scores}}
\label{tab:model_scores}
\end{table*}

\subsubsection{Zero-shot General Benchmarks}
Table \ref{tab:model_scores} also presents the comparison in zero-shot commonsense reasoning benchmark between our model and the two other model collections detailed in the previous section. As expected, \texttt{OnlySportsLM performs} the worst in all benchmarks, which is understandable given that it is only trained on sports-related text. For general-purpose models, we observe a positive correlation between their performance on sports domain tasks and their scores on commonsense reasoning benchmarks.

\subsubsection{Performance Across Training Steps}
In addition to cross-model comparison, we evaluate our model every 1000 checkpoints for \texttt{OnlySports Benchmark} and every 500 checkpoints for general benchmarks throughout the training process. This evaluation allows us to track the model's learning progression and identify any critical points or plateaus in performance 

\begin{figure*}[h]
  \includegraphics[width=0.48\linewidth]{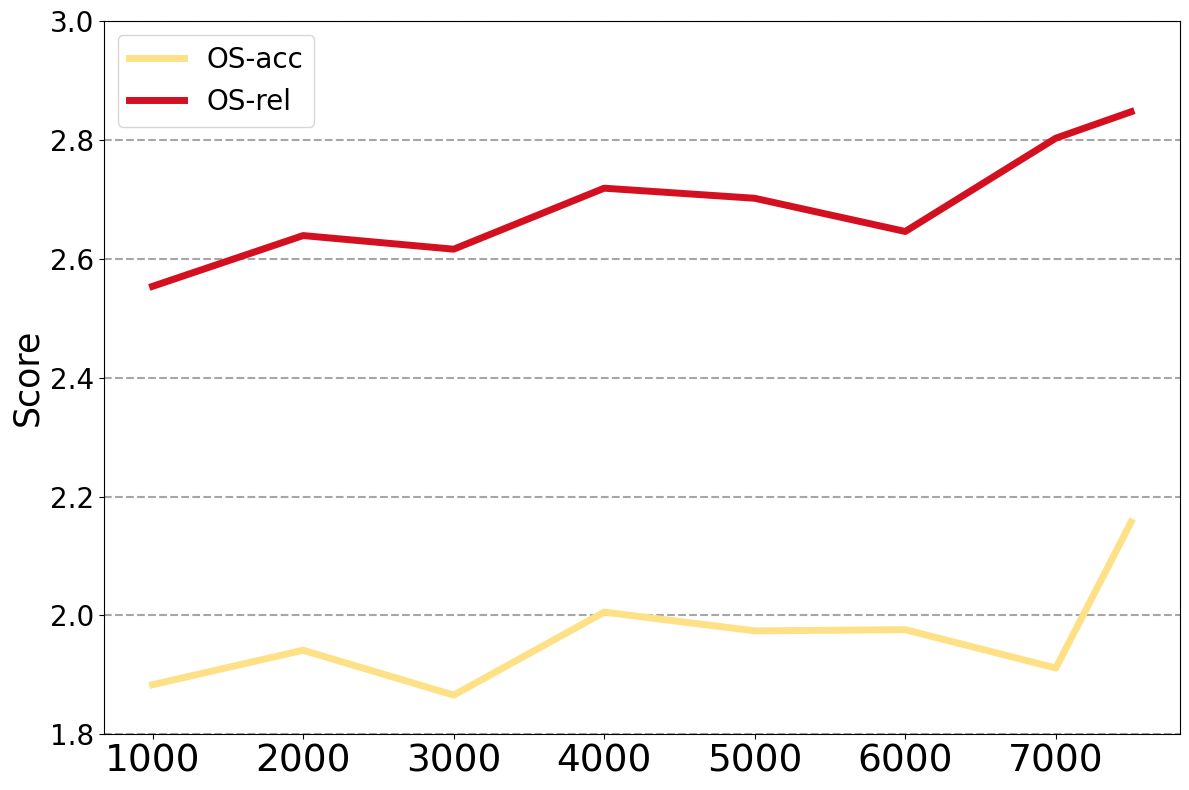} \hfill
  \includegraphics[width=0.48\linewidth]{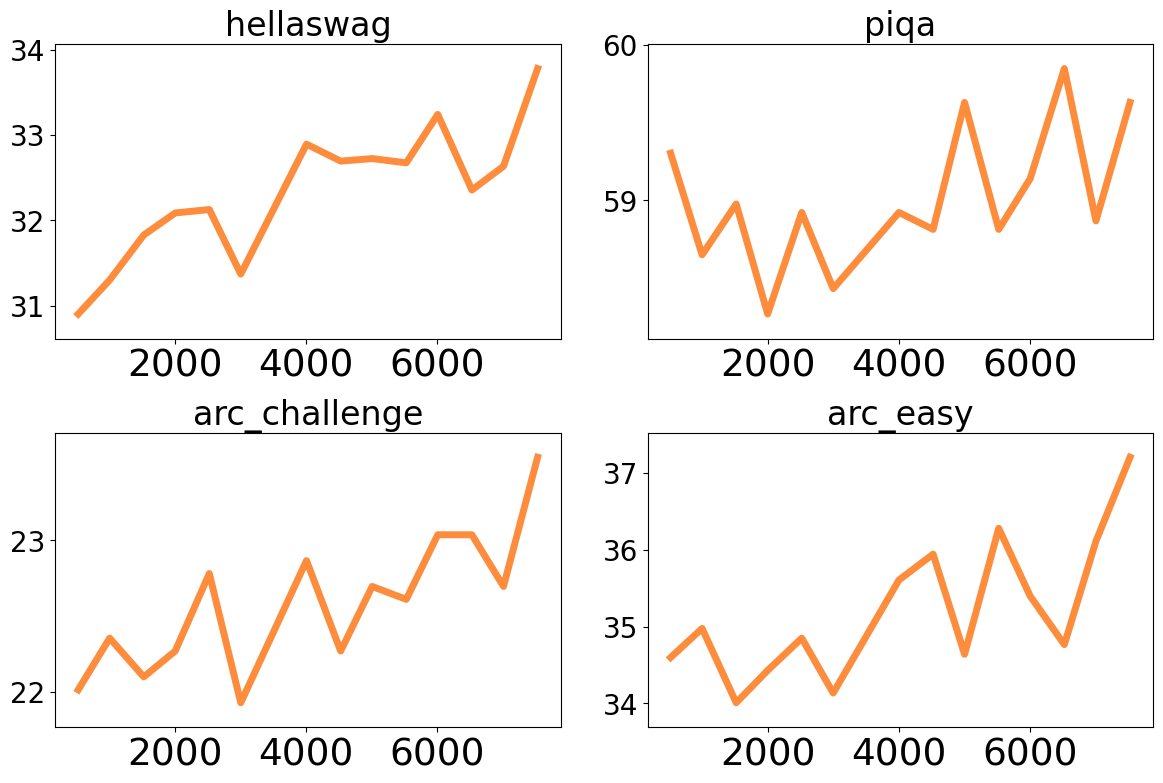}
  \caption {Evolution of \texttt{OnlySportsLM} performance across training steps. Left graph shows \texttt{OnlySports Benchmark} improving steadily. Right graphs display progress on general tasks, exhibiting upward trends despite fluctuations.}
  \label{fig:model_scores}
\end{figure*}
\noindent Figure \ref{fig:model_scores} presents our model's performance on various Benchmarks throughout the training process. We observe a consistent improvement in both \texttt{OS-acc} and \texttt{OS-rel} scores for \texttt{OnlySports Benchmarks} as training progressed. Surprisingly, we also notice performance improvements across all general benchmarks. This unexpected trend suggests that domain-specific training on sports-related text may enhance the model's general language understanding and commonsense reasoning capabilities. While the overall trend is positive, some fluctuations in performance were observed, particularly in the general benchmarks, which could be attributed to the complexities of the training process and the diverse nature of the evaluation tasks. 

\subsection{Future Work}
Building upon the promising results achieved with \texttt{OnlySportsLM}, future work will focus on exploring the model's full potential. We aim to complete training on the entire 600B token \texttt{OnlySports Dataset} when more funding is available, which may yield further improvements in both domain-specific and general language understanding. We also plan to explore instruction tuning techniques like instruction pre-training~\cite{cheng2024} and LAB~\cite{sudalairaj2024} to improve performance of our model. Additionally, we plan to investigate fine-tuning approaches for \texttt{OnlySportsLM}, potentially enhancing its performance on specific sports-related tasks. We are also interested in examining how domain improvements scale with increased model size, given that the performance of our model is comparable to other models with 1B parameters.

\section{Related Work}
Foundation models like GPT-4~\citep{openai2024gpt4technicalreport} and Llama 3~\cite{dubey2024llama3herdmodels} have demonstrated impressive performance on general-purpose language related tasks. These models are huge, with parameter ranges in hundreds of billions, and demand excessive computational resources to train. However, these general-purpose models fail to capture domain-specific nuances and context when generating content \citep{Yeung23,lin2024effectivenesslargelanguagemodels}. Though techniques like fine-tuning~\cite{zhang2023balancingspecializedgeneralskills,penedo2024finewebdatasetsdecantingweb} and prompt engineering~\cite{chen2024unleashingpotentialpromptengineering} can help in customization of general purpose LLMs for specific domains, the model size still remains an issue. 
\\

\noindent In parallel, efforts around developing domain-specific language models with models trained on in-domain data are also underway. Models like BloombergGPT~\cite{wu2023bloomberggptlargelanguagemodel} for finance, BioMedLM~\cite{bolton2024biomedlm27bparameterlanguage} for medical, and Galactica~\cite{taylor2022galacticalargelanguagemodel} for scientific research are LLMs trained on domain-specific data. These models also have billions of parameters and demand large-scale domain-specific dataset for training. The scale of training data and the computational cost has constrained wide-scale development of domain-specific LLMs. Further, excessive computational resources and energy requirement of such LLMs makes their deployment challenging on mobile devices thereby necessitating model compression through techniques like quantization~\cite{xiao23} and pruning~\cite{frantar23}.
\\

\noindent Recently~\citet{liu2024mobilellmoptimizingsubbillionparameter} developed MobileLLM, a sub-billion parameter family of LLMs achieving SOTA performance on standard language benchmarks. Through model architecture search they identified that deep and thin architectures achieve  better performance for compact LLMs. Within less than a month of the release of MobileLLM family, two new family of LLMs with sub-billion models, Qwen2~\citep{yang2024qwen2technicalreport} and SmolLM~\citep{benallal2024smollm}, were introduced. SmolLM-360M is claimed to beat performance of existing models with less than 500M parameters. The performance gains in SmolLM family are attributed to training using a well curated, high quality dataset. These work, though focused on developing general-purpose models, highlight the importance of data quality and model architecture optimization in developing high performing compact LLMs. Our \texttt{OnlySports} framework incorporates these insights when developing \texttt{OnlySportsLM}.
\\

\noindent In Sports domain, the only other model we found is SportsBERT~\cite{SportsBERT} which is a BERT \cite{devlin2019bertpretrainingdeepbidirectional} base language model trained on sports articles. However, there is no information on the dataset used to train the model and no evaluation of this model on sports-related language tasks.

\section{Conclusion}
This study focuses on optimizing sports domain language models with sub-billion parameters. Our findings demonstrate that for sports-related tasks, a carefully designed small model can outperform larger general-purpose models. By leveraging \texttt{OnlySports Dataset} and a carefully designed model architecture, we achieved significant improvements in sports knowledge generation and understanding. Our \texttt{OnlySportsLM}, a 196M parameter model, exhibits substantial advancements in sports-related text generation compared to previous state-of-the-art methods. The model's performance on \texttt{OnlySports Benchmark} underscores its effectiveness in continuing sports-related text. Furthermore, we demonstrate the potential of our approach in creating high-quality, domain-specific large datasets and evaluation methods. The \texttt{OnlySports Dataset} and \texttt{Benchmark} can provide valuable resources for future research in sports-related NLP tasks. Our study contributes to the ongoing research in developing efficient, domain-specific language models. While our approach shows promise in the sports domain, further investigation is needed to determine its adaptability to other specialized fields. We believe this work may offer insights that could be valuable for researchers exploring resource-efficient AI solutions across various domains.

\label{sec:bibtex}


\bibliography{acl_latex}

\begin{thebibliography}{28}
\providecommand{\natexlab}[1]{#1}

\bibitem[{Allal et~al.(2024)Allal, Lozhkov, and Bakouch}]{benallal2024smollm}
Loubna~Ben Allal, Anton Lozhkov, and Elie Bakouch. 2024.
\newblock \href {https://huggingface.co/blog/smollm} {Smollm - blazingly fast and remarkably powerful}.
\newblock Hugging Face Blog.
\newblock Published July 16, 2024.

\bibitem[{Au~Yeung et~al.(2023)Au~Yeung, Kraljevic, Luintel, Balston, Idowu, Dobson, and Teo}]{Yeung23}
Joshua Au~Yeung, Zeljko Kraljevic, Akish Luintel, Alfred Balston, Esther Idowu, Richard~J. Dobson, and James~T. Teo. 2023.
\newblock \href {https://doi.org/10.3389/fdgth.2023.1161098} {Ai chatbots not yet ready for clinical use}.
\newblock \emph{Frontiers in Digital Health}, 5.

\bibitem[{Bisk et~al.(2019)Bisk, Zellers, Bras, Gao, and Choi}]{bisk2019piqareasoningphysicalcommonsense}
Yonatan Bisk, Rowan Zellers, Ronan~Le Bras, Jianfeng Gao, and Yejin Choi. 2019.
\newblock \href {https://arxiv.org/abs/1911.11641} {Piqa: Reasoning about physical commonsense in natural language}.
\newblock \emph{Preprint}, arXiv:1911.11641.

\bibitem[{Bolton et~al.(2024)Bolton, Venigalla, Yasunaga, Hall, Xiong, Lee, Daneshjou, Frankle, Liang, Carbin, and Manning}]{bolton2024biomedlm27bparameterlanguage}
Elliot Bolton, Abhinav Venigalla, Michihiro Yasunaga, David Hall, Betty Xiong, Tony Lee, Roxana Daneshjou, Jonathan Frankle, Percy Liang, Michael Carbin, and Christopher~D. Manning. 2024.
\newblock \href {https://arxiv.org/abs/2403.18421} {Biomedlm: A 2.7b parameter language model trained on biomedical text}.
\newblock \emph{Preprint}, arXiv:2403.18421.

\bibitem[{Chen et~al.(2024)Chen, Zhang, Langrené, and Zhu}]{chen2024unleashingpotentialpromptengineering}
Banghao Chen, Zhaofeng Zhang, Nicolas Langrené, and Shengxin Zhu. 2024.
\newblock \href {https://arxiv.org/abs/2310.14735} {Unleashing the potential of prompt engineering in large language models: a comprehensive review}.
\newblock \emph{Preprint}, arXiv:2310.14735.

\bibitem[{Cheng et~al.(2024)Cheng, Gu, Huang, Bi, Huang, and Wei}]{cheng2024}
Daixuan Cheng, Yuxian Gu, Shaohan Huang, Junyu Bi, Minlie Huang, and Furu Wei. 2024.
\newblock \href {https://arxiv.org/abs/2406.14491} {Instruction pre-training: Language models are supervised multitask learners}.
\newblock \emph{Preprint}, arXiv:2406.14491.

\bibitem[{Clark et~al.(2018)Clark, Cowhey, Etzioni, Khot, Sabharwal, Schoenick, and Tafjord}]{clark2018thinksolvedquestionanswering}
Peter Clark, Isaac Cowhey, Oren Etzioni, Tushar Khot, Ashish Sabharwal, Carissa Schoenick, and Oyvind Tafjord. 2018.
\newblock \href {https://arxiv.org/abs/1803.05457} {Think you have solved question answering? try arc, the ai2 reasoning challenge}.
\newblock \emph{Preprint}, arXiv:1803.05457.

\bibitem[{Dean and Ghemawat(2008)}]{MapReduce}
Jeffrey Dean and Sanjay Ghemawat. 2008.
\newblock \href {https://doi.org/10.1145/1327452.1327492} {Mapreduce: simplified data processing on large clusters}.
\newblock \emph{Commun. ACM}, 51(1):107–113.

\bibitem[{Devlin et~al.(2019)Devlin, Chang, Lee, and Toutanova}]{devlin2019bertpretrainingdeepbidirectional}
Jacob Devlin, Ming-Wei Chang, Kenton Lee, and Kristina Toutanova. 2019.
\newblock \href {https://arxiv.org/abs/1810.04805} {Bert: Pre-training of deep bidirectional transformers for language understanding}.
\newblock \emph{Preprint}, arXiv:1810.04805.

\bibitem[{Dubey et~al.(2024)Dubey, Jauhri, Pandey, Kadian, Al-Dahle, Letman, Mathur, Schelten, Yang, Fan, Goyal, Hartshorn, Yang, Mitra, Sravankumar, Korenev, Hinsvark, Rao, Zhang, Rodriguez, Gregerson, Spataru, Roziere, Biron, Tang, Chern, Caucheteux, Nayak, Bi, Marra, McConnell, Keller, Touret, Wu, Wong, Ferrer, Nikolaidis, Allonsius, Song, Pintz, Livshits, Esiobu, Choudhary, Mahajan, Garcia-Olano, Perino, Hupkes, Lakomkin, AlBadawy, Lobanova, Dinan, Smith, Radenovic, Zhang, Synnaeve, Lee, Anderson, Nail, Mialon, Pang, Cucurell, Nguyen, Korevaar, Xu, Touvron, Zarov, Ibarra, Kloumann, Misra, Evtimov, Copet, Lee, Geffert, Vranes, Park, Mahadeokar, Shah, van~der Linde, Billock, Hong, Lee, Fu, Chi, Huang, Liu, Wang, Yu, Bitton, Spisak, Park, Rocca, Johnstun, Saxe, Jia, Alwala, Upasani, Plawiak, Li, Heafield, Stone, El-Arini, Iyer, Malik, Chiu, Bhalla, Rantala-Yeary, van~der Maaten, Chen, Tan, Jenkins, Martin, Madaan, Malo, Blecher, Landzaat, de~Oliveira, Muzzi, Pasupuleti, Singh, Paluri, Kardas, Oldham, Rita,
  Pavlova, Kambadur, Lewis, Si, Singh, Hassan, Goyal, Torabi, Bashlykov, Bogoychev, Chatterji, Duchenne, Çelebi, Alrassy, Zhang, Li, Vasic, Weng, Bhargava, Dubal, Krishnan, Koura, Xu, He, Dong, Srinivasan, Ganapathy, Calderer, Cabral, Stojnic, Raileanu, Girdhar, Patel, Sauvestre, Polidoro, Sumbaly, Taylor, Silva, Hou, Wang, Hosseini, Chennabasappa, Singh, Bell, Kim, Edunov, Nie, Narang, Raparthy, Shen, Wan, Bhosale, Zhang, Vandenhende, Batra, Whitman, Sootla, Collot, Gururangan, Borodinsky, Herman, Fowler, Sheasha, Georgiou, Scialom, Speckbacher, Mihaylov, Xiao, Karn, Goswami, Gupta, Ramanathan, Kerkez, Gonguet, Do, Vogeti, Petrovic, Chu, Xiong, Fu, Meers, Martinet, Wang, Tan, Xie, Jia, Wang, Goldschlag, Gaur, Babaei, Wen, Song, Zhang, Li, Mao, Coudert, Yan, Chen, Papakipos, Singh, Grattafiori, Jain, Kelsey, Shajnfeld, Gangidi, Victoria, Goldstand, Menon, Sharma, Boesenberg, Vaughan, Baevski, Feinstein, Kallet, Sangani, Yunus, Lupu, Alvarado, Caples, Gu, Ho, Poulton, Ryan, Ramchandani, Franco, Saraf,
  Chowdhury, Gabriel, Bharambe, Eisenman, Yazdan, James, Maurer, Leonhardi, Huang, Loyd, Paola, Paranjape, Liu, Wu, Ni, Hancock, Wasti, Spence, Stojkovic, Gamido, Montalvo, Parker, Burton, Mejia, Wang, Kim, Zhou, Hu, Chu, Cai, Tindal, Feichtenhofer, Civin, Beaty, Kreymer, Li, Wyatt, Adkins, Xu, Testuggine, David, Parikh, Liskovich, Foss, Wang, Le, Holland, Dowling, Jamil, Montgomery, Presani, Hahn, Wood, Brinkman, Arcaute, Dunbar, Smothers, Sun, Kreuk, Tian, Ozgenel, Caggioni, Guzmán, Kanayet, Seide, Florez, Schwarz, Badeer, Swee, Halpern, Thattai, Herman, Sizov, Guangyi, Zhang, Lakshminarayanan, Shojanazeri, Zou, Wang, Zha, Habeeb, Rudolph, Suk, Aspegren, Goldman, Damlaj, Molybog, Tufanov, Veliche, Gat, Weissman, Geboski, Kohli, Asher, Gaya, Marcus, Tang, Chan, Zhen, Reizenstein, Teboul, Zhong, Jin, Yang, Cummings, Carvill, Shepard, McPhie, Torres, Ginsburg, Wang, Wu, U, Saxena, Prasad, Khandelwal, Zand, Matosich, Veeraraghavan, Michelena, Li, Huang, Chawla, Lakhotia, Huang, Chen, Garg, A, Silva, Bell,
  Zhang, Guo, Yu, Moshkovich, Wehrstedt, Khabsa, Avalani, Bhatt, Tsimpoukelli, Mankus, Hasson, Lennie, Reso, Groshev, Naumov, Lathi, Keneally, Seltzer, Valko, Restrepo, Patel, Vyatskov, Samvelyan, Clark, Macey, Wang, Hermoso, Metanat, Rastegari, Bansal, Santhanam, Parks, White, Bawa, Singhal, Egebo, Usunier, Laptev, Dong, Zhang, Cheng, Chernoguz, Hart, Salpekar, Kalinli, Kent, Parekh, Saab, Balaji, Rittner, Bontrager, Roux, Dollar, Zvyagina, Ratanchandani, Yuvraj, Liang, Alao, Rodriguez, Ayub, Murthy, Nayani, Mitra, Li, Hogan, Battey, Wang, Maheswari, Howes, Rinott, Bondu, Datta, Chugh, Hunt, Dhillon, Sidorov, Pan, Verma, Yamamoto, Ramaswamy, Lindsay, Lindsay, Feng, Lin, Zha, Shankar, Zhang, Zhang, Wang, Agarwal, Sajuyigbe, Chintala, Max, Chen, Kehoe, Satterfield, Govindaprasad, Gupta, Cho, Virk, Subramanian, Choudhury, Goldman, Remez, Glaser, Best, Kohler, Robinson, Li, Zhang, Matthews, Chou, Shaked, Vontimitta, Ajayi, Montanez, Mohan, Kumar, Mangla, Albiero, Ionescu, Poenaru, Mihailescu, Ivanov, Li, Wang,
  Jiang, Bouaziz, Constable, Tang, Wang, Wu, Wang, Xia, Wu, Gao, Chen, Hu, Jia, Qi, Li, Zhang, Zhang, Adi, Nam, Yu, Wang, Hao, Qian, He, Rait, DeVito, Rosnbrick, Wen, Yang, and Zhao}]{dubey2024llama3herdmodels}
Abhimanyu Dubey, Abhinav Jauhri, Abhinav Pandey, Abhishek Kadian, Ahmad Al-Dahle, Aiesha Letman, Akhil Mathur, Alan Schelten, Amy Yang, Angela Fan, Anirudh Goyal, Anthony Hartshorn, Aobo Yang, Archi Mitra, Archie Sravankumar, Artem Korenev, Arthur Hinsvark, Arun Rao, Aston Zhang, Aurelien Rodriguez, Austen Gregerson, Ava Spataru, Baptiste Roziere, Bethany Biron, Binh Tang, Bobbie Chern, Charlotte Caucheteux, Chaya Nayak, Chloe Bi, Chris Marra, Chris McConnell, Christian Keller, Christophe Touret, Chunyang Wu, Corinne Wong, Cristian~Canton Ferrer, Cyrus Nikolaidis, Damien Allonsius, Daniel Song, Danielle Pintz, Danny Livshits, David Esiobu, Dhruv Choudhary, Dhruv Mahajan, Diego Garcia-Olano, Diego Perino, Dieuwke Hupkes, Egor Lakomkin, Ehab AlBadawy, Elina Lobanova, Emily Dinan, Eric~Michael Smith, Filip Radenovic, Frank Zhang, Gabriel Synnaeve, Gabrielle Lee, Georgia~Lewis Anderson, Graeme Nail, Gregoire Mialon, Guan Pang, Guillem Cucurell, Hailey Nguyen, Hannah Korevaar, Hu~Xu, Hugo Touvron, Iliyan Zarov,
  Imanol~Arrieta Ibarra, Isabel Kloumann, Ishan Misra, Ivan Evtimov, Jade Copet, Jaewon Lee, Jan Geffert, Jana Vranes, Jason Park, Jay Mahadeokar, Jeet Shah, Jelmer van~der Linde, Jennifer Billock, Jenny Hong, Jenya Lee, Jeremy Fu, Jianfeng Chi, Jianyu Huang, Jiawen Liu, Jie Wang, Jiecao Yu, Joanna Bitton, Joe Spisak, Jongsoo Park, Joseph Rocca, Joshua Johnstun, Joshua Saxe, Junteng Jia, Kalyan~Vasuden Alwala, Kartikeya Upasani, Kate Plawiak, Ke~Li, Kenneth Heafield, Kevin Stone, Khalid El-Arini, Krithika Iyer, Kshitiz Malik, Kuenley Chiu, Kunal Bhalla, Lauren Rantala-Yeary, Laurens van~der Maaten, Lawrence Chen, Liang Tan, Liz Jenkins, Louis Martin, Lovish Madaan, Lubo Malo, Lukas Blecher, Lukas Landzaat, Luke de~Oliveira, Madeline Muzzi, Mahesh Pasupuleti, Mannat Singh, Manohar Paluri, Marcin Kardas, Mathew Oldham, Mathieu Rita, Maya Pavlova, Melanie Kambadur, Mike Lewis, Min Si, Mitesh~Kumar Singh, Mona Hassan, Naman Goyal, Narjes Torabi, Nikolay Bashlykov, Nikolay Bogoychev, Niladri Chatterji, Olivier
  Duchenne, Onur Çelebi, Patrick Alrassy, Pengchuan Zhang, Pengwei Li, Petar Vasic, Peter Weng, Prajjwal Bhargava, Pratik Dubal, Praveen Krishnan, Punit~Singh Koura, Puxin Xu, Qing He, Qingxiao Dong, Ragavan Srinivasan, Raj Ganapathy, Ramon Calderer, Ricardo~Silveira Cabral, Robert Stojnic, Roberta Raileanu, Rohit Girdhar, Rohit Patel, Romain Sauvestre, Ronnie Polidoro, Roshan Sumbaly, Ross Taylor, Ruan Silva, Rui Hou, Rui Wang, Saghar Hosseini, Sahana Chennabasappa, Sanjay Singh, Sean Bell, Seohyun~Sonia Kim, Sergey Edunov, Shaoliang Nie, Sharan Narang, Sharath Raparthy, Sheng Shen, Shengye Wan, Shruti Bhosale, Shun Zhang, Simon Vandenhende, Soumya Batra, Spencer Whitman, Sten Sootla, Stephane Collot, Suchin Gururangan, Sydney Borodinsky, Tamar Herman, Tara Fowler, Tarek Sheasha, Thomas Georgiou, Thomas Scialom, Tobias Speckbacher, Todor Mihaylov, Tong Xiao, Ujjwal Karn, Vedanuj Goswami, Vibhor Gupta, Vignesh Ramanathan, Viktor Kerkez, Vincent Gonguet, Virginie Do, Vish Vogeti, Vladan Petrovic, Weiwei Chu,
  Wenhan Xiong, Wenyin Fu, Whitney Meers, Xavier Martinet, Xiaodong Wang, Xiaoqing~Ellen Tan, Xinfeng Xie, Xuchao Jia, Xuewei Wang, Yaelle Goldschlag, Yashesh Gaur, Yasmine Babaei, Yi~Wen, Yiwen Song, Yuchen Zhang, Yue Li, Yuning Mao, Zacharie~Delpierre Coudert, Zheng Yan, Zhengxing Chen, Zoe Papakipos, Aaditya Singh, Aaron Grattafiori, Abha Jain, Adam Kelsey, Adam Shajnfeld, Adithya Gangidi, Adolfo Victoria, Ahuva Goldstand, Ajay Menon, Ajay Sharma, Alex Boesenberg, Alex Vaughan, Alexei Baevski, Allie Feinstein, Amanda Kallet, Amit Sangani, Anam Yunus, Andrei Lupu, Andres Alvarado, Andrew Caples, Andrew Gu, Andrew Ho, Andrew Poulton, Andrew Ryan, Ankit Ramchandani, Annie Franco, Aparajita Saraf, Arkabandhu Chowdhury, Ashley Gabriel, Ashwin Bharambe, Assaf Eisenman, Azadeh Yazdan, Beau James, Ben Maurer, Benjamin Leonhardi, Bernie Huang, Beth Loyd, Beto~De Paola, Bhargavi Paranjape, Bing Liu, Bo~Wu, Boyu Ni, Braden Hancock, Bram Wasti, Brandon Spence, Brani Stojkovic, Brian Gamido, Britt Montalvo, Carl
  Parker, Carly Burton, Catalina Mejia, Changhan Wang, Changkyu Kim, Chao Zhou, Chester Hu, Ching-Hsiang Chu, Chris Cai, Chris Tindal, Christoph Feichtenhofer, Damon Civin, Dana Beaty, Daniel Kreymer, Daniel Li, Danny Wyatt, David Adkins, David Xu, Davide Testuggine, Delia David, Devi Parikh, Diana Liskovich, Didem Foss, Dingkang Wang, Duc Le, Dustin Holland, Edward Dowling, Eissa Jamil, Elaine Montgomery, Eleonora Presani, Emily Hahn, Emily Wood, Erik Brinkman, Esteban Arcaute, Evan Dunbar, Evan Smothers, Fei Sun, Felix Kreuk, Feng Tian, Firat Ozgenel, Francesco Caggioni, Francisco Guzmán, Frank Kanayet, Frank Seide, Gabriela~Medina Florez, Gabriella Schwarz, Gada Badeer, Georgia Swee, Gil Halpern, Govind Thattai, Grant Herman, Grigory Sizov, Guangyi, Zhang, Guna Lakshminarayanan, Hamid Shojanazeri, Han Zou, Hannah Wang, Hanwen Zha, Haroun Habeeb, Harrison Rudolph, Helen Suk, Henry Aspegren, Hunter Goldman, Ibrahim Damlaj, Igor Molybog, Igor Tufanov, Irina-Elena Veliche, Itai Gat, Jake Weissman, James
  Geboski, James Kohli, Japhet Asher, Jean-Baptiste Gaya, Jeff Marcus, Jeff Tang, Jennifer Chan, Jenny Zhen, Jeremy Reizenstein, Jeremy Teboul, Jessica Zhong, Jian Jin, Jingyi Yang, Joe Cummings, Jon Carvill, Jon Shepard, Jonathan McPhie, Jonathan Torres, Josh Ginsburg, Junjie Wang, Kai Wu, Kam~Hou U, Karan Saxena, Karthik Prasad, Kartikay Khandelwal, Katayoun Zand, Kathy Matosich, Kaushik Veeraraghavan, Kelly Michelena, Keqian Li, Kun Huang, Kunal Chawla, Kushal Lakhotia, Kyle Huang, Lailin Chen, Lakshya Garg, Lavender A, Leandro Silva, Lee Bell, Lei Zhang, Liangpeng Guo, Licheng Yu, Liron Moshkovich, Luca Wehrstedt, Madian Khabsa, Manav Avalani, Manish Bhatt, Maria Tsimpoukelli, Martynas Mankus, Matan Hasson, Matthew Lennie, Matthias Reso, Maxim Groshev, Maxim Naumov, Maya Lathi, Meghan Keneally, Michael~L. Seltzer, Michal Valko, Michelle Restrepo, Mihir Patel, Mik Vyatskov, Mikayel Samvelyan, Mike Clark, Mike Macey, Mike Wang, Miquel~Jubert Hermoso, Mo~Metanat, Mohammad Rastegari, Munish Bansal, Nandhini
  Santhanam, Natascha Parks, Natasha White, Navyata Bawa, Nayan Singhal, Nick Egebo, Nicolas Usunier, Nikolay~Pavlovich Laptev, Ning Dong, Ning Zhang, Norman Cheng, Oleg Chernoguz, Olivia Hart, Omkar Salpekar, Ozlem Kalinli, Parkin Kent, Parth Parekh, Paul Saab, Pavan Balaji, Pedro Rittner, Philip Bontrager, Pierre Roux, Piotr Dollar, Polina Zvyagina, Prashant Ratanchandani, Pritish Yuvraj, Qian Liang, Rachad Alao, Rachel Rodriguez, Rafi Ayub, Raghotham Murthy, Raghu Nayani, Rahul Mitra, Raymond Li, Rebekkah Hogan, Robin Battey, Rocky Wang, Rohan Maheswari, Russ Howes, Ruty Rinott, Sai~Jayesh Bondu, Samyak Datta, Sara Chugh, Sara Hunt, Sargun Dhillon, Sasha Sidorov, Satadru Pan, Saurabh Verma, Seiji Yamamoto, Sharadh Ramaswamy, Shaun Lindsay, Shaun Lindsay, Sheng Feng, Shenghao Lin, Shengxin~Cindy Zha, Shiva Shankar, Shuqiang Zhang, Shuqiang Zhang, Sinong Wang, Sneha Agarwal, Soji Sajuyigbe, Soumith Chintala, Stephanie Max, Stephen Chen, Steve Kehoe, Steve Satterfield, Sudarshan Govindaprasad, Sumit Gupta,
  Sungmin Cho, Sunny Virk, Suraj Subramanian, Sy~Choudhury, Sydney Goldman, Tal Remez, Tamar Glaser, Tamara Best, Thilo Kohler, Thomas Robinson, Tianhe Li, Tianjun Zhang, Tim Matthews, Timothy Chou, Tzook Shaked, Varun Vontimitta, Victoria Ajayi, Victoria Montanez, Vijai Mohan, Vinay~Satish Kumar, Vishal Mangla, Vítor Albiero, Vlad Ionescu, Vlad Poenaru, Vlad~Tiberiu Mihailescu, Vladimir Ivanov, Wei Li, Wenchen Wang, Wenwen Jiang, Wes Bouaziz, Will Constable, Xiaocheng Tang, Xiaofang Wang, Xiaojian Wu, Xiaolan Wang, Xide Xia, Xilun Wu, Xinbo Gao, Yanjun Chen, Ye~Hu, Ye~Jia, Ye~Qi, Yenda Li, Yilin Zhang, Ying Zhang, Yossi Adi, Youngjin Nam, Yu, Wang, Yuchen Hao, Yundi Qian, Yuzi He, Zach Rait, Zachary DeVito, Zef Rosnbrick, Zhaoduo Wen, Zhenyu Yang, and Zhiwei Zhao. 2024.
\newblock \href {https://arxiv.org/abs/2407.21783} {The llama 3 herd of models}.
\newblock \emph{Preprint}, arXiv:2407.21783.

\bibitem[{Frantar and Alistarh(2023)}]{frantar23}
Elias Frantar and Dan Alistarh. 2023.
\newblock Sparsegpt: massive language models can be accurately pruned in one-shot.
\newblock In \emph{Proceedings of the 40th International Conference on Machine Learning}, ICML'23. JMLR.org.

\bibitem[{Lin et~al.(2024)Lin, Chen, Hu, Zhang, Wan, Wei, Xu, Wang, and Gu}]{lin2024effectivenesslargelanguagemodels}
Yalan Lin, Meng Chen, Yuhan Hu, Hongyu Zhang, Chengcheng Wan, Zhao Wei, Yong Xu, Juhong Wang, and Xiaodong Gu. 2024.
\newblock \href {https://arxiv.org/abs/2312.01639} {On the effectiveness of large language models in domain-specific code generation}.
\newblock \emph{Preprint}, arXiv:2312.01639.

\bibitem[{Liu et~al.(2024)Liu, Zhao, Iandola, Lai, Tian, Fedorov, Xiong, Chang, Shi, Krishnamoorthi, Lai, and Chandra}]{liu2024mobilellmoptimizingsubbillionparameter}
Zechun Liu, Changsheng Zhao, Forrest Iandola, Chen Lai, Yuandong Tian, Igor Fedorov, Yunyang Xiong, Ernie Chang, Yangyang Shi, Raghuraman Krishnamoorthi, Liangzhen Lai, and Vikas Chandra. 2024.
\newblock \href {https://arxiv.org/abs/2402.14905} {Mobilellm: Optimizing sub-billion parameter language models for on-device use cases}.
\newblock \emph{Preprint}, arXiv:2402.14905.

\bibitem[{Loshchilov and Hutter(2019)}]{loshchilov2019decoupledweightdecayregularization}
Ilya Loshchilov and Frank Hutter. 2019.
\newblock \href {https://arxiv.org/abs/1711.05101} {Decoupled weight decay regularization}.
\newblock \emph{Preprint}, arXiv:1711.05101.

\bibitem[{Merrick(2024)}]{merrick2024embeddingclusteringdataimprove}
Luke Merrick. 2024.
\newblock \href {https://arxiv.org/abs/2407.18887} {Embedding and clustering your data can improve contrastive pretraining}.
\newblock \emph{Preprint}, arXiv:2407.18887.

\bibitem[{Microsoft()}]{SportsBERT}
Microsoft.
\newblock "sportsbert".
\newblock \url{https://huggingface.co/microsoft/SportsBERT}.
\newblock Accessed: 2024-08-27.

\bibitem[{Minaee et~al.(2024)Minaee, Mikolov, Nikzad, Chenaghlu, Socher, Amatriain, and Gao}]{minaee2024largelanguagemodelssurvey}
Shervin Minaee, Tomas Mikolov, Narjes Nikzad, Meysam Chenaghlu, Richard Socher, Xavier Amatriain, and Jianfeng Gao. 2024.
\newblock \href {https://arxiv.org/abs/2402.06196} {Large language models: A survey}.
\newblock \emph{Preprint}, arXiv:2402.06196.

\bibitem[{OpenAI et~al.(2024)OpenAI, Achiam, Adler, Agarwal, Ahmad, Akkaya, Aleman, Almeida, Altenschmidt, Altman, Anadkat, Avila, Babuschkin, Balaji, Balcom, Baltescu, Bao, Bavarian, Belgum, Bello, Berdine, Bernadett-Shapiro, Berner, Bogdonoff, Boiko, Boyd, Brakman, Brockman, Brooks, Brundage, Button, Cai, Campbell, Cann, Carey, Carlson, Carmichael, Chan, Chang, Chantzis, Chen, Chen, Chen, Chen, Chen, Chess, Cho, Chu, Chung, Cummings, Currier, Dai, Decareaux, Degry, Deutsch, Deville, Dhar, Dohan, Dowling, Dunning, Ecoffet, Eleti, Eloundou, Farhi, Fedus, Felix, Fishman, Forte, Fulford, Gao, Georges, Gibson, Goel, Gogineni, Goh, Gontijo-Lopes, Gordon, Grafstein, Gray, Greene, Gross, Gu, Guo, Hallacy, Han, Harris, He, Heaton, Heidecke, Hesse, Hickey, Hickey, Hoeschele, Houghton, Hsu, Hu, Hu, Huizinga, Jain, Jain, Jang, Jiang, Jiang, Jin, Jin, Jomoto, Jonn, Jun, Kaftan, Łukasz Kaiser, Kamali, Kanitscheider, Keskar, Khan, Kilpatrick, Kim, Kim, Kim, Kirchner, Kiros, Knight, Kokotajlo, Łukasz Kondraciuk,
  Kondrich, Konstantinidis, Kosic, Krueger, Kuo, Lampe, Lan, Lee, Leike, Leung, Levy, Li, Lim, Lin, Lin, Litwin, Lopez, Lowe, Lue, Makanju, Malfacini, Manning, Markov, Markovski, Martin, Mayer, Mayne, McGrew, McKinney, McLeavey, McMillan, McNeil, Medina, Mehta, Menick, Metz, Mishchenko, Mishkin, Monaco, Morikawa, Mossing, Mu, Murati, Murk, Mély, Nair, Nakano, Nayak, Neelakantan, Ngo, Noh, Ouyang, O'Keefe, Pachocki, Paino, Palermo, Pantuliano, Parascandolo, Parish, Parparita, Passos, Pavlov, Peng, Perelman, de~Avila Belbute~Peres, Petrov, de~Oliveira~Pinto, Michael, Pokorny, Pokrass, Pong, Powell, Power, Power, Proehl, Puri, Radford, Rae, Ramesh, Raymond, Real, Rimbach, Ross, Rotsted, Roussez, Ryder, Saltarelli, Sanders, Santurkar, Sastry, Schmidt, Schnurr, Schulman, Selsam, Sheppard, Sherbakov, Shieh, Shoker, Shyam, Sidor, Sigler, Simens, Sitkin, Slama, Sohl, Sokolowsky, Song, Staudacher, Such, Summers, Sutskever, Tang, Tezak, Thompson, Tillet, Tootoonchian, Tseng, Tuggle, Turley, Tworek, Uribe, Vallone,
  Vijayvergiya, Voss, Wainwright, Wang, Wang, Wang, Ward, Wei, Weinmann, Welihinda, Welinder, Weng, Weng, Wiethoff, Willner, Winter, Wolrich, Wong, Workman, Wu, Wu, Wu, Xiao, Xu, Yoo, Yu, Yuan, Zaremba, Zellers, Zhang, Zhang, Zhao, Zheng, Zhuang, Zhuk, and Zoph}]{openai2024gpt4technicalreport}
OpenAI, Josh Achiam, Steven Adler, Sandhini Agarwal, Lama Ahmad, Ilge Akkaya, Florencia~Leoni Aleman, Diogo Almeida, Janko Altenschmidt, Sam Altman, Shyamal Anadkat, Red Avila, Igor Babuschkin, Suchir Balaji, Valerie Balcom, Paul Baltescu, Haiming Bao, Mohammad Bavarian, Jeff Belgum, Irwan Bello, Jake Berdine, Gabriel Bernadett-Shapiro, Christopher Berner, Lenny Bogdonoff, Oleg Boiko, Madelaine Boyd, Anna-Luisa Brakman, Greg Brockman, Tim Brooks, Miles Brundage, Kevin Button, Trevor Cai, Rosie Campbell, Andrew Cann, Brittany Carey, Chelsea Carlson, Rory Carmichael, Brooke Chan, Che Chang, Fotis Chantzis, Derek Chen, Sully Chen, Ruby Chen, Jason Chen, Mark Chen, Ben Chess, Chester Cho, Casey Chu, Hyung~Won Chung, Dave Cummings, Jeremiah Currier, Yunxing Dai, Cory Decareaux, Thomas Degry, Noah Deutsch, Damien Deville, Arka Dhar, David Dohan, Steve Dowling, Sheila Dunning, Adrien Ecoffet, Atty Eleti, Tyna Eloundou, David Farhi, Liam Fedus, Niko Felix, Simón~Posada Fishman, Juston Forte, Isabella Fulford, Leo
  Gao, Elie Georges, Christian Gibson, Vik Goel, Tarun Gogineni, Gabriel Goh, Rapha Gontijo-Lopes, Jonathan Gordon, Morgan Grafstein, Scott Gray, Ryan Greene, Joshua Gross, Shixiang~Shane Gu, Yufei Guo, Chris Hallacy, Jesse Han, Jeff Harris, Yuchen He, Mike Heaton, Johannes Heidecke, Chris Hesse, Alan Hickey, Wade Hickey, Peter Hoeschele, Brandon Houghton, Kenny Hsu, Shengli Hu, Xin Hu, Joost Huizinga, Shantanu Jain, Shawn Jain, Joanne Jang, Angela Jiang, Roger Jiang, Haozhun Jin, Denny Jin, Shino Jomoto, Billie Jonn, Heewoo Jun, Tomer Kaftan, Łukasz Kaiser, Ali Kamali, Ingmar Kanitscheider, Nitish~Shirish Keskar, Tabarak Khan, Logan Kilpatrick, Jong~Wook Kim, Christina Kim, Yongjik Kim, Jan~Hendrik Kirchner, Jamie Kiros, Matt Knight, Daniel Kokotajlo, Łukasz Kondraciuk, Andrew Kondrich, Aris Konstantinidis, Kyle Kosic, Gretchen Krueger, Vishal Kuo, Michael Lampe, Ikai Lan, Teddy Lee, Jan Leike, Jade Leung, Daniel Levy, Chak~Ming Li, Rachel Lim, Molly Lin, Stephanie Lin, Mateusz Litwin, Theresa Lopez, Ryan
  Lowe, Patricia Lue, Anna Makanju, Kim Malfacini, Sam Manning, Todor Markov, Yaniv Markovski, Bianca Martin, Katie Mayer, Andrew Mayne, Bob McGrew, Scott~Mayer McKinney, Christine McLeavey, Paul McMillan, Jake McNeil, David Medina, Aalok Mehta, Jacob Menick, Luke Metz, Andrey Mishchenko, Pamela Mishkin, Vinnie Monaco, Evan Morikawa, Daniel Mossing, Tong Mu, Mira Murati, Oleg Murk, David Mély, Ashvin Nair, Reiichiro Nakano, Rajeev Nayak, Arvind Neelakantan, Richard Ngo, Hyeonwoo Noh, Long Ouyang, Cullen O'Keefe, Jakub Pachocki, Alex Paino, Joe Palermo, Ashley Pantuliano, Giambattista Parascandolo, Joel Parish, Emy Parparita, Alex Passos, Mikhail Pavlov, Andrew Peng, Adam Perelman, Filipe de~Avila Belbute~Peres, Michael Petrov, Henrique~Ponde de~Oliveira~Pinto, Michael, Pokorny, Michelle Pokrass, Vitchyr~H. Pong, Tolly Powell, Alethea Power, Boris Power, Elizabeth Proehl, Raul Puri, Alec Radford, Jack Rae, Aditya Ramesh, Cameron Raymond, Francis Real, Kendra Rimbach, Carl Ross, Bob Rotsted, Henri Roussez,
  Nick Ryder, Mario Saltarelli, Ted Sanders, Shibani Santurkar, Girish Sastry, Heather Schmidt, David Schnurr, John Schulman, Daniel Selsam, Kyla Sheppard, Toki Sherbakov, Jessica Shieh, Sarah Shoker, Pranav Shyam, Szymon Sidor, Eric Sigler, Maddie Simens, Jordan Sitkin, Katarina Slama, Ian Sohl, Benjamin Sokolowsky, Yang Song, Natalie Staudacher, Felipe~Petroski Such, Natalie Summers, Ilya Sutskever, Jie Tang, Nikolas Tezak, Madeleine~B. Thompson, Phil Tillet, Amin Tootoonchian, Elizabeth Tseng, Preston Tuggle, Nick Turley, Jerry Tworek, Juan Felipe~Cerón Uribe, Andrea Vallone, Arun Vijayvergiya, Chelsea Voss, Carroll Wainwright, Justin~Jay Wang, Alvin Wang, Ben Wang, Jonathan Ward, Jason Wei, CJ~Weinmann, Akila Welihinda, Peter Welinder, Jiayi Weng, Lilian Weng, Matt Wiethoff, Dave Willner, Clemens Winter, Samuel Wolrich, Hannah Wong, Lauren Workman, Sherwin Wu, Jeff Wu, Michael Wu, Kai Xiao, Tao Xu, Sarah Yoo, Kevin Yu, Qiming Yuan, Wojciech Zaremba, Rowan Zellers, Chong Zhang, Marvin Zhang, Shengjia
  Zhao, Tianhao Zheng, Juntang Zhuang, William Zhuk, and Barret Zoph. 2024.
\newblock \href {https://arxiv.org/abs/2303.08774} {Gpt-4 technical report}.
\newblock \emph{Preprint}, arXiv:2303.08774.

\bibitem[{Penedo et~al.(2024)Penedo, Kydlíček, allal, Lozhkov, Mitchell, Raffel, Werra, and Wolf}]{penedo2024finewebdatasetsdecantingweb}
Guilherme Penedo, Hynek Kydlíček, Loubna~Ben allal, Anton Lozhkov, Margaret Mitchell, Colin Raffel, Leandro~Von Werra, and Thomas Wolf. 2024.
\newblock \href {https://arxiv.org/abs/2406.17557} {The fineweb datasets: Decanting the web for the finest text data at scale}.
\newblock \emph{Preprint}, arXiv:2406.17557.

\bibitem[{Peng et~al.(2024)Peng, Goldstein, Anthony, Albalak, Alcaide, Biderman, Cheah, Du, Ferdinan, Hou, Kazienko, GV, Kocoń, Koptyra, Krishna, au2, Muennighoff, Obeid, Saito, Song, Tu, Woźniak, Zhang, Zhao, Zhao, Zhou, Zhu, and Zhu}]{peng2024eaglefinchrwkvmatrixvalued}
Bo~Peng, Daniel Goldstein, Quentin Anthony, Alon Albalak, Eric Alcaide, Stella Biderman, Eugene Cheah, Xingjian Du, Teddy Ferdinan, Haowen Hou, Przemysław Kazienko, Kranthi~Kiran GV, Jan Kocoń, Bartłomiej Koptyra, Satyapriya Krishna, Ronald McClelland~Jr. au2, Niklas Muennighoff, Fares Obeid, Atsushi Saito, Guangyu Song, Haoqin Tu, Stanisław Woźniak, Ruichong Zhang, Bingchen Zhao, Qihang Zhao, Peng Zhou, Jian Zhu, and Rui-Jie Zhu. 2024.
\newblock \href {https://arxiv.org/abs/2404.05892} {Eagle and finch: Rwkv with matrix-valued states and dynamic recurrence}.
\newblock \emph{Preprint}, arXiv:2404.05892.

\bibitem[{Sudalairaj et~al.(2024)Sudalairaj, Bhandwaldar, Pareja, Xu, Cox, and Srivastava}]{sudalairaj2024}
Shivchander Sudalairaj, Abhishek Bhandwaldar, Aldo Pareja, Kai Xu, David~D. Cox, and Akash Srivastava. 2024.
\newblock \href {https://arxiv.org/abs/2403.01081} {Lab: Large-scale alignment for chatbots}.
\newblock \emph{Preprint}, arXiv:2403.01081.

\bibitem[{Taylor et~al.(2022)Taylor, Kardas, Cucurull, Scialom, Hartshorn, Saravia, Poulton, Kerkez, and Stojnic}]{taylor2022galacticalargelanguagemodel}
Ross Taylor, Marcin Kardas, Guillem Cucurull, Thomas Scialom, Anthony Hartshorn, Elvis Saravia, Andrew Poulton, Viktor Kerkez, and Robert Stojnic. 2022.
\newblock \href {https://arxiv.org/abs/2211.09085} {Galactica: A large language model for science}.
\newblock \emph{Preprint}, arXiv:2211.09085.

\bibitem[{Wu et~al.(2023)Wu, Irsoy, Lu, Dabravolski, Dredze, Gehrmann, Kambadur, Rosenberg, and Mann}]{wu2023bloomberggptlargelanguagemodel}
Shijie Wu, Ozan Irsoy, Steven Lu, Vadim Dabravolski, Mark Dredze, Sebastian Gehrmann, Prabhanjan Kambadur, David Rosenberg, and Gideon Mann. 2023.
\newblock \href {https://arxiv.org/abs/2303.17564} {Bloomberggpt: A large language model for finance}.
\newblock \emph{Preprint}, arXiv:2303.17564.

\bibitem[{Xiao et~al.(2023)Xiao, Lin, Seznec, Wu, Demouth, and Han}]{xiao23}
Guangxuan Xiao, Ji~Lin, Mickael Seznec, Hao Wu, Julien Demouth, and Song Han. 2023.
\newblock Smoothquant: accurate and efficient post-training quantization for large language models.
\newblock In \emph{Proceedings of the 40th International Conference on Machine Learning}, ICML'23. JMLR.org.

\bibitem[{Yang et~al.(2024)Yang, Yang, Hui, Zheng, Yu, Zhou, Li, Li, Liu, Huang, Dong, Wei, Lin, Tang, Wang, Yang, Tu, Zhang, Ma, Yang, Xu, Zhou, Bai, He, Lin, Dang, Lu, Chen, Yang, Li, Xue, Ni, Zhang, Wang, Peng, Men, Gao, Lin, Wang, Bai, Tan, Zhu, Li, Liu, Ge, Deng, Zhou, Ren, Zhang, Wei, Ren, Liu, Fan, Yao, Zhang, Wan, Chu, Liu, Cui, Zhang, Guo, and Fan}]{yang2024qwen2technicalreport}
An~Yang, Baosong Yang, Binyuan Hui, Bo~Zheng, Bowen Yu, Chang Zhou, Chengpeng Li, Chengyuan Li, Dayiheng Liu, Fei Huang, Guanting Dong, Haoran Wei, Huan Lin, Jialong Tang, Jialin Wang, Jian Yang, Jianhong Tu, Jianwei Zhang, Jianxin Ma, Jianxin Yang, Jin Xu, Jingren Zhou, Jinze Bai, Jinzheng He, Junyang Lin, Kai Dang, Keming Lu, Keqin Chen, Kexin Yang, Mei Li, Mingfeng Xue, Na~Ni, Pei Zhang, Peng Wang, Ru~Peng, Rui Men, Ruize Gao, Runji Lin, Shijie Wang, Shuai Bai, Sinan Tan, Tianhang Zhu, Tianhao Li, Tianyu Liu, Wenbin Ge, Xiaodong Deng, Xiaohuan Zhou, Xingzhang Ren, Xinyu Zhang, Xipin Wei, Xuancheng Ren, Xuejing Liu, Yang Fan, Yang Yao, Yichang Zhang, Yu~Wan, Yunfei Chu, Yuqiong Liu, Zeyu Cui, Zhenru Zhang, Zhifang Guo, and Zhihao Fan. 2024.
\newblock \href {https://arxiv.org/abs/2407.10671} {Qwen2 technical report}.
\newblock \emph{Preprint}, arXiv:2407.10671.

\bibitem[{Zellers et~al.(2019)Zellers, Holtzman, Bisk, Farhadi, and Choi}]{zellers2019hellaswagmachinereallyfinish}
Rowan Zellers, Ari Holtzman, Yonatan Bisk, Ali Farhadi, and Yejin Choi. 2019.
\newblock \href {https://arxiv.org/abs/1905.07830} {Hellaswag: Can a machine really finish your sentence?}
\newblock \emph{Preprint}, arXiv:1905.07830.

\bibitem[{Zhang et~al.(2023)Zhang, Zheng, Tang, Sun, Ma, Bu, Zhou, and Zhao}]{zhang2023balancingspecializedgeneralskills}
Zheng Zhang, Chen Zheng, Da~Tang, Ke~Sun, Yukun Ma, Yingtong Bu, Xun Zhou, and Liang Zhao. 2023.
\newblock \href {https://arxiv.org/abs/2310.04945} {Balancing specialized and general skills in llms: The impact of modern tuning and data strategy}.
\newblock \emph{Preprint}, arXiv:2310.04945.

\bibitem[{Zheng et~al.(2024)Zheng, Chiang, Sheng, Zhuang, Wu, Zhuang, Lin, Li, Li, Xing, Zhang, Gonzalez, and Stoica}]{Zheng2023}
Lianmin Zheng, Wei-Lin Chiang, Ying Sheng, Siyuan Zhuang, Zhanghao Wu, Yonghao Zhuang, Zi~Lin, Zhuohan Li, Dacheng Li, Eric~P. Xing, Hao Zhang, Joseph~E. Gonzalez, and Ion Stoica. 2024.
\newblock Judging llm-as-a-judge with mt-bench and chatbot arena.
\newblock In \emph{Proceedings of the 37th International Conference on Neural Information Processing Systems}, NIPS '23.

\end{thebibliography}

\appendix

\section{Appendix}
\subsection{Evaluation Criteria}
This appendix provides detailed grading rubrics for the two main evaluation criteria used in the \texttt{OnlySports Benchmark}: Accuracy and Factuality (OS-acc), and Continuity and Relevancy (OS-rel). These rubrics were provided to the GPT-4 and Claude 3.5 Sonnet models as part of their system messages when acting as evaluators. \{num\} specified the number of responses that will be in the prompt for evaluation.

\subsubsection{Accuracy and Factuality}
Prompt for evaluating accuracy and factuality:
\textit{You are a sports expert assigned to grade language models' generation performance on general sports-related text according to the provided rubric. 
1 prompt and \{num\} responses will be presented, all attempting to complete the same given prompt. Each response is separated by [SEP] and limited to 80 tokens.}
\\

\noindent \textit{Evaluate responses using the following rubric for "Accuracy and Factuality":}
\textit{\\
"1": "Mostly inaccurate, significant factual errors.", \\
"2": "Partially accurate, mix of correct and incorrect information.",\\ "3": "Mostly accurate, minor factual errors.", \\"4": "Highly accurate, negligible errors.",\\ "5": "Fully accurate and factually impeccable."}
\\

\noindent \textit{When evaluating, only consider the accuracy and factuality in the context of the given prompt.
Score these generated responses on a scale of 1-5. Only output the scores! Output scores in the following format: 'X, X, X, X, X...', where X is a number between 1 and 5.}

\subsubsection{Continuity and Relevance}
Prompt for evaluating continuity and relevance:
\textit{You are a sports expert assigned to grade language models' generation performance on general sports-related text according to the provided rubric. 
1 prompt and \{num\} responses will be presented, all attempting to complete the same given prompt. Each response is started by [SEP] and limited to 80 tokens.}
\\

\noindent \textit{Evaluate responses using the following rubric for "Continuity and Relevance":}
\textit{\\
"1": "Poor continuation, diverges significantly from the prompt's context or topic.",
    \\
    "2": "Weak continuation, maintains some elements of the prompt but introduces unrelated content.",
    \\
    "3": "Adequate continuation, generally follows the prompt's direction with some minor deviations.",
    \\
    "4": "Strong continuation, closely follows the prompt's context and style with minimal inconsistencies.",
    \\
    "5": "Excellent continuation, seamlessly extends the prompt's narrative, context, and style."}
\\

\noindent \textit{When evaluating, only consider how well the generated text continues or extends the given prompt in terms of context, topic, and style.
Score these generated responses on a scale of 1-5. Only output the scores! Output scores in the following format: 'X, X, X, X, X...', where X is a number between 1 and 5.}

\subsection{Example Prompts and Responses for OnlySports Benchmark Scoring}
\label{sec:appendix_examples}

This appendix provides concrete examples of prompts and \texttt{OnlySportsLM} responses to illustrate the scoring criteria used in the \texttt{OnlySports Benchmark}. We present four distinct scenarios that demonstrate various combinations of Accuracy and Factuality (OS-acc) and Continuity and Relevancy (OS-rel) scores:

1. High Accuracy, High Relevancy
2. High Accuracy, Low Relevancy
3. Low Accuracy, High Relevancy
4. Low Accuracy, Low Relevancy

For each scenario, we provide the original prompt, response, and a brief explanation of the scoring rationale. These examples are designed to offer clarity on how the evaluation criteria were applied and to aid in the reproducibility of our results.

\noindent 1. Prompt: \textit{Named after the famed coach, the Vince Lombardi Trophy is awarded to the team that wins}

Response: \textit{the Super Bowl. The trophy is awarded to the team that wins the Super Bowl, and the trophy is named after the coach who won the most Super Bowls. The trophy is named after Vince Lombardi, who won the Super Bowl with the Green Bay Packers in 1967.
The }
\\

\noindent This response accurately states that the Vince Lombardi Trophy is awarded to the Super Bowl winner. It correctly mentions Vince Lombardi and his connection to the Green Bay Packers. The information is directly relevant to the prompt and provides additional context about the trophy's history.
\\

\noindent 2. Prompt: \textit{With flippers and a flutter kick, scuba divers use a modified version of the freestyle stroke to}

Response: \textit{keep their hands and feet on the ground. The dives are performed in a single-piece, high-tech suit. The diver is required to wear a mask and a cap. In the 2012 London Olympics, the dives were judged by a panel of experts. In 2013, the International Olympic Committee (IOC) banned the dives.}
\\

\noindent While the response contains accurate information about diving suits and Olympic judging, it's not relevant to the prompt about scuba diving techniques. The mention of flippers and freestyle stroke is ignored, making the response off-topic despite its factual correctness.
\\

\noindent 3. Prompt: \textit{"As the Lakers head toward another challenging season, they are hopeful about LeBron James's"}

Response:\textit{' decision to return to the team. James, who will be a free agent this summer, has not made a decision on his future."I'm just happy that he's back," Lakers coach Frank Vogel said. "I think he's a big part of our team. He's a big part of our team. I think he's going to be a big part of our team'}
\\

\noindent The response is relevant to the Lakers and LeBron James's future with the team, matching the prompt's topic. However, it contains inaccuracies: LeBron James was not a free agent at the time specified, and Frank Vogel was no longer the Lakers' coach. The content is on-topic but factually incorrect.
\\

\noindent 4. Prompt: \textit{Hockey legend Wayne Gretzky holds the NHL record for the most points scored in a single season at}

Response: \textit{100 points.
NHL players are not the only ones who have the ability to score goals. Some of the most popular sports in the world include soccer, basketball, baseball, football, and hockey. In addition to the games themselves, the sport of hockey is also a popular pastime for many people. There are many ways to play hockey, and the rules of the game}
\\

\noindent This response is both inaccurate and irrelevant. The stated record of 100 points is incorrect (Gretzky's record is 215 points). Most of the response diverges into general sports discussion unrelated to Gretzky's NHL record, making it largely irrelevant to the prompt.

\subsection{Model Scores from Claude and GPT}
\label{sub:scores}
In this section, we present the raw score from the two state-of-the-art LLMs: Claude 3.5 and GPT-4o. The table below (Table 4) showcases the scores for each model across OS-rel (Relevance) and OS-acc (Accuracy). These scores provide insight into how well each model performs in generating sports-related content, with higher scores indicating better performance. We observe that Claude 3.5 Sonnet generally gives higher scores than GPT4o, using the same prompt.

\begin{table}[h]
\centering
\begin{tabular}{l|cc|cc}
\hline
\textbf{Model} & \multicolumn{2}{c|}{\textbf{Claude 3.5}} & \multicolumn{2}{c}{\textbf{GPT-4o}} \\
 & \textbf{OS-rel} & \textbf{OS-acc} & \textbf{OS-rel} & \textbf{OS-acc} \\

\hline
OnlySportsLM & 3.19 & 2.38 & 2.50 & 1.94 \\
Qwen2-0.5B & 2.34 & 1.93 & 1.82 & 1.36 \\
Qwen2-1.5B & 3.23 & 2.73 & 2.68 & 1.93 \\
SmolLM-135M & 2.25 & 1.96 & 1.66 & 1.41 \\
SmolLM-360M & 2.23 & 1.91 & 1.82 & 1.50\\
SmolLM-1.7B & 2.97 & 2.55 & 2.48 & 1.97 \\
\hline
\end{tabular}
\caption{Performance scores for different language models across two evaluators}
\label{tab:model-scores}
\end{table}
\end{document}